\documentclass{article}
\usepackage{graphicx} 

\usepackage{amsmath}
\usepackage{amssymb}
\usepackage{amsthm}
\usepackage{graphicx}
\usepackage[colorlinks=true, allcolors=blue]{hyperref}
\usepackage{subcaption}
\usepackage{cite}
\usepackage{enumitem}
\usepackage{algorithm, algorithmic}

\usepackage{adjustbox}

\usepackage{pifont,tabularx,adjustbox,booktabs}

\newcolumntype{R}[2]{%
>{\adjustbox{angle=#1,lap=\width-(#2)}\bgroup}%
    l%
    <{\egroup}%
}

\newcommand*\rot{\multicolumn{1}{R{30}{1em}}}

\usepackage{titlesec}
\usepackage[titletoc]{appendix}
\usepackage{varwidth}

\def\ol{\overline}
\def\ul{\underline}

\def\wt{\tilde}

\def\st {\fct{s.t.}}

\def\cont {\fct{co}}

\def\trial {\fct{trial}}
\def\best {\fct{best}}

\def\Rz{\mathbb{R}}

\def\fct#1{\mathop{\rm #1}}	
 
\def\x {{\bf X}}

\usepackage{tikz}
\usetikzlibrary{matrix,shapes,arrows,positioning,chains}

\usepackage[colorlinks=true, allcolors=blue]{hyperref}
\usepackage{orcidlink} 

\begin{document}

\begin{center}

{\Large \bf Machine Learning Algorithms for Improving Black Box Optimization Solvers} 

{\large \bf Morteza Kimiaei$^{\ast}$\orcidlink{0000-0002-7973-3770}}

\centerline{\sl Fakult\"at f\"ur Mathematik, Universit\"at Wien}
\centerline{\sl Oskar-Morgenstern-Platz 1, A-1090 Wien, Austria}
\centerline{\sl Email: \href{mailto:kimiaeim83@univie.ac.at}{kimiaeim83@univie.ac.at}}
\centerline{\sl WWW: \url{http://www.mat.univie.ac.at/~kimiaei}}
{\small ($^\ast$Corresponding Author)}
\vspace{0.5cm}

{\large \bf Vyacheslav Kungurtsev\orcidlink{0000-0003-2229-8824}}
\centerline{\sl Department of Computer Science, Czech Technical University}
\centerline{\sl Karlovo Namesti 13, 121 35 Prague
2, Czech Republic}
{\sl Email: \href{mailto:vyacheslav.kungurtsev@fel.cvut.cz}{vyacheslav.kungurtsev@fel.cvut.cz}}

\end{center}

\begin{sloppypar}
{\bf Abstract.} Black-box optimization (BBO) addresses problems where objectives are accessible only through costly queries without gradients or explicit structure.  Classical derivative-free methods---line search, direct search, and model-based solvers such as Bayesian optimization---form the backbone of BBO, yet often struggle in high-dimensional, noisy, or mixed-integer settings. Recent advances use machine learning (ML) and reinforcement learning (RL) to enhance BBO: ML provides expressive surrogates, adaptive updates, meta-learning-based algorithm selection, and generative models, while the RL--BBO interface includes dynamic operator configuration, robust policy search, and meta-optimization across tasks. This paper surveys these developments, covering representative algorithms such as the modular model-based optimization framework ({\tt mlrMBO}), zeroth-order adaptive momentum methods ({\tt ZO-AdaMM}), automated BBO ({\tt ABBO}), distributed block-wise optimization ({\tt DiBB}), partition-based Bayesian optimization ({\tt SPBOpt}), the transformer-based optimizer ({\tt B2Opt}), diffusion-model-based BBO, surrogate-assisted RL for differential evolution ({\tt Surr-RLDE}), robust BBO ({\tt RBO}), coordinate-ascent MORE with step size adaptation ({\tt CAS-MORE}), log-barrier stochastic gradient descent ({\tt LB-SGD}), policy improvement with black-box ({\tt PIBB}), and offline Q-learning with Mamba backbones ({\tt Q-Mamba}).  We further review benchmark efforts such as the NeurIPS 2020 BBO Challenge and the {\tt MetaBox} framework. A controlled numerical study further confirms that ML-based candidate screening can improve the evaluation-budget performance of direct search and differential evolution, while the benefit remains solver-dependent.
\end{sloppypar}

\clearpage



\begin{sloppypar}

\section{Introduction}

Machine learning (ML) techniques have recently gained significant attention for their ability to model complex relationships and to enhance the performance of optimization algorithms across a wide range of application domains. In Black-Box Optimization (BBO), where explicit structural information about the objective or constraints is unavailable, ML offers principled mechanisms for improving the efficiency, robustness, and scalability of classical solvers.

\subsection{Background and Motivation}

BBO problems frequently arise in science and engineering when the objective can only be evaluated through costly simulations, experiments, or function calls, and neither gradients nor structural information are available. These problems are inherently challenging because the search space may be high-dimensional, multimodal, and nonconvex, and evaluations may be noisy or computationally expensive. As a result, exact optimization strategies are often impractical, leading practitioners to rely on heuristic or stochastic algorithms such as evolutionary strategies, Bayesian optimization, and surrogate-assisted methods.  

\subsection{Problem Setting and Research Question} 

In this context, ML models offer principled approaches for accelerating the optimization process. They can leverage previously collected evaluations to construct predictive surrogates, guide exploration--exploitation trade-offs, or adaptively tune algorithmic hyperparameters. Reinforcement Learning (RL) approaches, for example, can be used to learn search policies that dynamically allocate resources across candidate solutions. In contrast, supervised learning methods can extract patterns from historical runs to inform initialization or sampling strategies.  

We first define the set of simple bounds 
\begin{equation}\label{e.box}
\x:=\{x\in\Rz^n\mid \ul x\le x \le \ol x\}   \ \ \mbox{with $\ul x,\ol x\in \Rz^n$ ($\ul x<\ol x$)}
\end{equation} 
on variables $x\in\Rz^n$ (called the {\bf box}).  
Then, the continuous BBO problem can be expressed in the same formal way as
\begin{equation}\label{e.CNLP}
\begin{array}{ll}
\min & f(x)\\
\st  & x\in C_{\cont},\\
\end{array}
\end{equation}
where $f:C_{\cont} \subseteq \x  \to \Rz$ is a real-valued (possibly nonconvex) black box objective function defined on the {\bf continuous nonlinear feasible set}
\begin{equation}\label{e.contSet}
C_{\cont}:=\{x \in \x \mid g(x)=0, \ \ h(x)\le 0\}.
\end{equation}
Here, $g(x)=(g_1(x),\ldots,g_m(x))$ and $h(x)=(h_1(x),\ldots,h_p(x))$ represent equality and inequality constraints, respectively, both treated as black box functions accessible only through evaluations.  For a positive integer $q$, we set
\[
[q] := \{1,2,\ldots,q\}.
\]
We will use this notation throughout the paper.

In line with prior surveys and the book of Audet and Hare~\cite{AudH}, our formal problem setting and primary discussion focus on {\bf continuous} BBO methods. Integer and mixed-integer extensions are mentioned when they are intrinsic capabilities of the surveyed frameworks, but they are not reviewed systematically in this work.

From this formal definition, two central research questions arise: (i) how can learning-based models, particularly ML and RL, be systematically integrated into the BBO framework to improve sample efficiency, scalability, and robustness; and (ii) to what extent do such methods generalize across tasks and domains under limited evaluation budgets.  This survey aims to address these questions by analyzing how data-driven components augment or replace traditional search mechanisms within BBO, and by identifying common methodological patterns that unify classical Derivative-Free Optimization (DFO) approaches with modern ML and RL paradigms.

\subsection{Challenges in BBO}

Despite significant progress, solving large-scale BBO problems remains a fundamental challenge due to their high-dimensional, nonconvex, and often noisy nature.  
Unlike problems with explicit analytical structure, BBO tasks rely solely on function evaluations that may be stochastic, expensive, or discontinuous, making gradient information unavailable or unreliable.  
As the dimensionality or evaluation cost increases, the search space expands exponentially, and the number of required queries often becomes prohibitive, particularly when simulation or experimental evaluations dominate the computational budget.  

These challenges are particularly severe in domains such as automated design, robotics, and hyperparameter tuning, where each evaluation may require minutes to hours of computation.  
Even with recent advances in surrogate modeling, parallel sampling, and meta-learning, state-of-the-art solvers struggle to maintain efficiency and robustness across diverse problem types.  
For instance, while Bayesian optimization and evolution strategies perform well in moderate dimensions, they deteriorate rapidly beyond a few hundred variables, and diffusion- or transformer-based BBO frameworks face scalability bottlenecks during training and inference.  

As noted by recent benchmarking efforts (e.g., {\tt MetaBox}~\cite{ma2023metabox}, {\tt MetaBBO-RL}~\cite{ma2023metabox}), industrial-scale BBO applications, such as aerodynamic shape optimization, photonics design, or multi-agent policy search, remain challenging for general-purpose solvers, with performance depending on structure and computational resources.  Practitioners thus increasingly rely on hybrid or approximate approaches that combine surrogate learning, parallel sampling, and domain-specific priors to balance computational feasibility with solution quality.

\subsection{Scope and Methodology}

This survey focuses on BBO/DFO frameworks that integrate ML and RL to enhance solver components such as surrogate modeling, adaptive step-size control, operator configuration, and meta-policy design. The learning-related literature considered here spans \textbf{2012 to 2025}, with earlier foundational work also cited, covering the evolution from foundational DFO and stochastic search algorithms (e.g., \cite{LarMW,RioS}) to modern learning-augmented and neural generative BBO paradigms (e.g., \cite{li2025b2opt,li2024diffusionbbo,ma2025qmamba}).  

Sources were identified through targeted searches in \textit{Scopus}, \textit{Web of Science}, and \textit{arXiv} using key terms such as 
\textbf{ML for BBO},  
\textbf{RL for DFO},  
\textbf{meta-BBO}, and  
\textbf{neural or diffusion models for BBO.}  Inclusion criteria required that studies  
(i) employ ML or RL within a BBO or DFO context where objective or constraint functions are opaque and gradients unavailable,  (ii) provide either empirical validation on benchmark suites 
(e.g., {\tt COCO}/{\tt BBOB}\footnote{\url{https://numbbo.github.io/coco/}}
or continuous-control tasks in
{\tt MuJoCo}\footnote{\url{https://mujoco.org/}}) 
or theoretical insight into convergence and robustness, and  
(iii) be peer-reviewed or recognized preprints with reproducible implementations or benchmark access (e.g., {\tt MetaBox}~\cite{ma2023metabox}, {\tt BBO Challenge 2020}~\cite{candelieri2020bboc}).  

This search strategy provides broad coverage from classical BBO foundations, such as direct search, line search, and trust-region methods, to the latest ML- and RL-enhanced frameworks, including adaptive momentum solvers ({\tt ZO-AdaMM}~\cite{chen2019zo}), surrogate-based model optimizers ({\tt mlrMBO}~\cite{bischl2018mlrmbo}), and algorithm selection and meta-optimization ({\tt ABBO}~\cite{meunier2022abbo}, {\tt Surr-RLDE}~\cite{ma2025surr-rlde}, {\tt Q-Mamba}~\cite{ma2025qmamba}). Together, these studies trace the methodological progression of BBO from gradient-free search toward intelligent, data-driven optimization.  

In this paper, we review developments in \textbf{ML- and RL-enhanced BBO}, emphasizing how learning mechanisms extend the capabilities of traditional DFO solvers. Our analysis begins by outlining the classical BBO taxonomy, polling-, surrogate-, and local-approximation-based methods, and proceeds to explore how ML contributes through surrogate modeling, optimizer-inspired updates, meta-learning, and generative modeling. RL-based enhancements are then examined for their roles in robustness, operator adaptation, and meta-policy learning.  Section~\ref{sec:NNs-BBO} forms the core of this survey, highlighting the methodological shift from algorithmic heuristics to \textbf{learning-augmented BBO}, which integrates predictive modeling and decision-making into the optimization process itself.

\subsection{Contributions}

The main contributions of this paper are as follows:
\begin{itemize}
  \item In Section \ref{sec:BBO}, we provide a unified taxonomy of classical BBO methods
        (line search, direct search, model-based, Bayesian optimization) and
        position recent ML and RL advances 
        as enhancements that extend these foundations.
  \item In Section \ref{sec:NNs-BBO}, we review recent ML and RL approaches designed to complement
        \textbf{classical BBO solvers}. Our emphasis is on their integration into
        \textbf{inexact solution methods}---procedures that do not guarantee global
        optimality but strive to deliver high-quality solutions under strict time or
        evaluation budgets. We begin with a brief overview of traditional BBO heuristics
        and then discuss how data-driven methods are being employed to extend or transform
        these strategies.
\begin{itemize}
  \item In Subsection \ref{sec:MLenhance}, we survey major \textbf{ML-enhanced BBO} frameworks, including surrogate-based approaches such as the modular model-based optimization framework ({\tt mlrMBO} \cite{bischl2018mlrmbo}), search partitioning for Bayesian optimization ({\tt SPBOpt} \cite{sazanovich2021spbopt}), and trust-region derivative-free optimization 
({\tt DFO-TR} \cite{ghanbari2017dfotr}); optimizer-inspired updates such as the zeroth-order adaptive momentum method ({\tt ZO-AdaMM} \cite{chen2019zo}); benchmarking-informed portfolios such as {\tt ABBO} \cite{meunier2022abbo}, and the related distributed framework {\tt DiBB} \cite{cuccu2022dibb}; and generative optimizers such as the transformer-based optimizer ({\tt B2Opt} \cite{li2025b2opt}) and diffusion-model-based BBO (\cite{li2024diffusionbbo}).
  \item In Subsection \ref{sec:RLenhance}, we review \textbf{the RL--BBO interface}, including robust policy search through 
        robust black-box optimization 
        ({\tt RBO} \cite{choromanski2019rbo}), 
        log-barrier stochastic gradient descent 
        ({\tt LB-SGD} \cite{usmanova2024logsafe}), 
        and coordinate-ascent MORE with step size adaptation 
        ({\tt CAS-MORE} \cite{huettenrauch2024more}); 
        dynamic operator configuration via surrogate-assisted RL for 
        differential evolution 
        ({\tt Surr-RLDE} \cite{ma2025surr-rlde}); 
        policy search connections through policy improvement with black-box updates 
        ({\tt PIBB} \cite{stulp2012pibb}); 
        and offline meta-optimization with decomposed Q-learning using a Mamba backbone 
        ({\tt Q-Mamba} \cite{ma2025qmamba}).
\end{itemize}
\item In Section \ref{sec:survey}, we highlight \textbf{benchmarking and reproducibility efforts}, including the 
NeurIPS 2020 BBO Challenge \cite{candelieri2020bboc}, the {\tt MetaBox} framework \cite{ma2023metabox}, and recent {\tt MetaBBO-RL} evaluation studies \cite{ma2023metabox}, establishing standardized protocols for fair comparison.

\item In Section \ref{sec:num-ml}, we complement the survey with reproducible numerical experiments showing improved success rates and final solution quality for the learning-enhanced direct search and differential evolution variants under matched evaluation budgets.

\item We emphasize the role of \textbf{continuous BBO} as a central application domain, where ML and RL provide practical efficiency and robustness beyond classical heuristics.
\end{itemize}

This paper is accompanied by supplemental material (\texttt{SuppMat.pdf}) and the numerical code and data. 

The supplement provides the pseudocode for all BBO methods and discusses how ML and RL can enhance these approaches. It also reviews evolutionary population-based techniques and zero-order methods with approximate gradients, along with background material on ML and RL. Finally, it summarizes existing benchmarking results on the use of ML and RL within BBO methods. To help readers navigate the many algorithms discussed in this survey and their detailed pseudocode in the supplemental material, Table~\ref{t.Map} provides an algorithm map linking each method to its corresponding section and supplemental algorithm number.

\begin{table}[h]
\centering
\caption{Map of Algorithms Between Main Paper and Supplemental Material}\label{t.Map}
\resizebox{\textwidth}{!}{%
\begin{tabular}{lll}
\toprule
\textbf{Algorithm / Method} & \textbf{Main Paper Section} & \textbf{Supplement Algorithm} \\
\midrule
\texttt{SBO} & Section~\ref{sec:SBO} & Algorithm 1 \\
\texttt{DS} & Section~\ref{sec:DS} & Algorithm 2 \\
\texttt{LS-dd} & Section~\ref{sec:DS} & Algorithm 3 \\
\texttt{LS+dd} & Section~\ref{sec:LS+dd} & Algorithm 4 \\
\texttt{TR} & Section~\ref{sec:LS+dd} & Algorithm 1 \\
\texttt{Generic-BBO-Workflow} & Section~\ref{sec:GWBBO} & Algorithm 5 \\
\texttt{mlrMBO} & Section~\ref{sec:mlrMBO} & Algorithm 6 \\
\texttt{ZO-AdaMM} & Section~\ref{sec:ZO-AdaMM} & Algorithm 7 \\
\texttt{ABBO} & Section~\ref{sec:ABBO} & Algorithm 8 \\
\texttt{DFO-TR} & Section~\ref{sec:DFO-TR} & Algorithm 9 \\
\texttt{SPBOpt} & Section~\ref{sec:SPBOpt} & Algorithm 10 \\
\texttt{B2Opt} & Section~\ref{sec:B2Opt} & Algorithm 11 \\
\texttt{DiffBBO} & Section~\ref{sec:DiffBBO} & Algorithm 12 \\
\texttt{DiBB} & Section~\ref{sec:DiBB} & Algorithm 13 \\
\texttt{Surr-RLDE} & Section~\ref{sec:Surr-RLDE} & Algorithm 14 \\
\texttt{RBO} & Section~\ref{sec:RBO} & Algorithm 15 \\
\texttt{CAS-MORE} & Section~\ref{sec:CAS-MORE} & Algorithm 16 \\
\texttt{LB-SGD} & Section~\ref{sec:LB-SGD} & Algorithm 17 \\
\texttt{PI2}/\texttt{PIBB} & Section~\ref{sec:PIBB} & Algorithm 18 \\
\texttt{Q-Mamba} & Section~\ref{sec:Q-Mamba} & Algorithm 19 \\
\texttt{BBO-Challenge-2020} & Section~\ref{sec:BBO-C-2020} & Algorithm 20 \\
\texttt{MetaBox} & Section~\ref{sec:MetaBBO} & Algorithm 21 \\
\texttt{MetaBBO-RL-Benchmarking} & Section~\ref{sec:MetaBBO-RL} & Algorithm 22 \\
\bottomrule
\end{tabular}}
\end{table}

\section{BBO Methods}\label{sec:BBO}

BBO methods form the algorithmic foundation for tackling problems where the objective function and constraints are available only through evaluations, with no derivative or structural information. In this section, we survey the main families of classical derivative-free optimization (DFO) solvers, polling-based, surrogate-based, and local-approximation-based, as well as stochastic and evolutionary algorithms. These methods provide the groundwork upon which modern ML- and RL-enhanced approaches to BBO are built.

DFO encompasses a class of optimization algorithms designed to minimize or maximize an objective function without requiring gradient information. These methods are essential in scenarios where the objective function is non-differentiable, noisy, or computationally expensive, such as hyperparameter tuning, BBO, or model calibration in ML.

DFO algorithms and their applications were discussed in the books of Audet \& Hare \cite{AudH} and Conn et al. \cite{ConSV} and the survey paper of Larson et al. \cite{LarMW}. The numerical behavior of integer and mixed-integer DFO solvers was investigated in the survey of Ploskas \& Sahinidis \cite{Ploskas2021}, while the numerical behavior of noiseless continuous DFO algorithms was investigated in the papers by Mor{\'{e}} \& Wild \cite{MorW}, Rios \& Sahinidis \cite{RioS}, Kimiaei et al. \cite{SSDFO}, and Kimiaei \& Neumaier \cite{VRBBO}, and of noisy continuous DFO algorithms was investigated in the recent papers of Kimiaei \cite{VRDFON} and Kimiaei \& Neumaier \cite{MADFO}.

For organizational purposes, we discuss three overlapping groups
\cite{AudH,ConSV,LarMW}: \textbf{polling-based methods},
\textbf{surrogate-based methods}, and
\textbf{local-approximation-based methods}. The distinction
emphasizes search mechanisms rather than mutually exclusive
classes. Bayesian optimization uses probabilistic surrogates and
acquisition functions; trust-region DFO constructs local models. 

Line-search ({\tt LS}) methods can be divided into two classes depending on whether approximate gradient information is used. The first {\tt LS} class corresponds to {\tt LS+dd}, while the second {\tt LS} class corresponds to {\tt LS-dd}.

{\tt LS+dd} employs an approximated gradient, and the {\tt LS} condition can take the form of the approximate Armijo, Goldstein, or Wolfe conditions. 
This class naturally falls under the category of {\bf local-approximation-based methods}, since it relies on constructing first-order models to guide search. 
In noisy DFO, however, the efficiency of these methods is significantly reduced, as the approximate gradient may be inaccurate. 
Examples include the Armijo line search, which uses backtracking interpolation to enforce the Armijo condition; the Wolfe line search, which enforces both the Armijo and curvature conditions through interpolation and extrapolation; and the Goldstein line search, which extends Armijo with a two-sided condition. 
Recently, Neumaier et al.~\cite{CLS} proposed an improved Goldstein method that combines interpolation and extrapolation more efficiently. Representative solvers in this class include {\tt SSDFO} \cite{SSDFO} and {\tt FMINUNC} \cite{FMINUNC}, both based on approximate Wolfe conditions. 

{\tt LS-dd} replaces directional derivative scaled by a step size with a forcing function (positive and non-decreasing, e.g., $c_1\alpha^2$ with the tuning parameter $0<c_1<1$ and the step size $\alpha>0$) as the line-search condition. This class fits into the category of {\bf polling-based methods}, since it relies solely on function evaluations along search directions without constructing explicit gradient approximations. 
Representative solvers in this category include {\tt DFLINT} \cite{DFLINT} (integer deterministic {\tt LS}), {\tt DFLBOX} \cite{DFLBOX} and {\tt DFNDFL} \cite{DFNDFL} (integer and continuous deterministic {\tt LS}), {\tt VRBBO} \cite{VRBBO} and {\tt VRDFON} \cite{VRDFON} (continuous randomized {\tt LS}), and {\tt SDBOX} \cite{SDBOX} (continuous deterministic {\tt LS}). Applicability and scalability depend on the algorithm variant and problem structure. Here, {\tt FMINUNC} is considered with numerical gradients; it is a gradient-based solver rather than a dedicated derivative-free algorithm.

 In summary, classical DFO methods span a wide spectrum, from polling-based algorithms that rely solely on function values, to surrogate-based approaches constructing global models, and local-approximation methods that use first- or second-order information within trust-region or line-search frameworks. This taxonomy provides a unified view of the methodological foundations upon which more advanced, ML-enhanced BBO techniques are built.

\subsection{Surrogate-Based Methods}\label{sec:SBO}

The goal of surrogate-based optimization ({\tt SBO}) is to construct and refine an inexpensive model of the objective function, which may be stochastic, noisy, or otherwise expensive to evaluate. By iteratively updating the surrogate with new evaluations, {\tt SBO} methods guide the search toward promising regions of the domain while reducing the number of costly function calls  \cite{Rasmussen2006GPML,Garnett2023BO}. Algorithm 1 in {\tt SuppMat.pdf} provides a generic framework for {\tt SBO} applied to the CNLP problem \eqref{e.CNLP}.

{\tt SBO} constructs various surrogate models such as Gaussian processes, radial basis functions, and polynomial regression. The initial model can be constructed by selecting sampling points, and then the models are updated using new evaluations, including unsuccessful trials when useful. {\tt pySOT} \cite{pySOT}, {\tt Dakota} \cite{Dakota}, {\tt SPLINE} \cite{SPLINE}, and {\tt MISO} \cite{Mller2015} are the four {\tt SBO} solvers. {\tt pySOT} is a flexible surrogate optimization framework supporting various surrogates and methods such as Gaussian processes, Kriging, and radial basis functions for global optimization and handling expensive black-box functions,  {\tt Dakota} is a multilevel parallel object-oriented framework using a wide range of optimization algorithms (e.g., surrogate-based optimization, trust-region algorithms, and evolutionary algorithms) for high-dimensional and multi-fidelity optimization problems. The method denoted by {\tt SPLINE} uses radial basis function interpolation for global optimization. {\tt MISO} is a surrogate optimization framework for expensive mixed-integer black-box problems.

{\bf Gaussian Process ({\tt GP}) Surrogates.} A {\tt GP} surrogate defines a prior over functions 
$f:\Rz^d \to \Rz$ such that for any finite set $\{x_i\}_{i=1}^N$, the vector 
$(f(x_1),\dots,f(x_N))$ follows a multivariate Gaussian distribution.  
A {\tt GP} is specified by a mean function $m_0(x)$ and a kernel (covariance function) 
$k(x,x')$, giving predictive mean and variance
\[
\begin{aligned}
\mu(x) &= m_0(x)+k(x,X)(K+\sigma_n^2I)^{-1}(y-m_0(X)),\\
\sigma^2(x) &= k(x,x)-k(x,X)(K+\sigma_n^2I)^{-1}k(X,x).
\end{aligned}
\]
where $K$ is the kernel matrix on the training inputs $X$, $y$ are observed outputs, 
$m_0(X)$ contains the prior means, and $\sigma_n^2$ is the observation-noise variance.  
{\tt GP}s provide posterior predictions and model-conditional uncertainty estimates, 
which are crucial for acquisition functions in Bayesian optimization.

{\bf Radial Basis Function ({\tt RBF}) Surrogates.}
An \texttt{RBF} surrogate models the objective as a weighted sum of radially symmetric kernels centered at previously evaluated points:
\[
\hat f(x) \;=\; \sum_{i=1}^N w_i \,\phi(\|x - x_i\|),
\]
where $\{x_i\}_{i=1}^N \subset \Rz^d$ are the $N$ sampled design points, $w_i \in \Rz$ are interpolation weights chosen to fit $f(x_i)$ at these points, $\phi: \Rz^+ \to \Rz$ is a radial kernel, commonly $\phi(r) = \exp(-\gamma r^2)$ with shape parameter $\gamma>0$ (Gaussian {\tt RBF}), or alternatives such as multiquadric or thin-plate splines. Because \texttt{RBF} surrogates interpolate known evaluations and are smooth, 
they are widely used in trust-region DFO  to approximate the objective within the local trust-region.

{\bf Bayesian optimization ({\tt BO}).} The model-based {\tt BO} methods construct a probabilistic model (often a {\tt GP}) to approximate the nonlinear objective function and combine uncertainty to balance exploration and exploitation. A key component of {\tt BO} is the {\bf acquisition function}, which guides 
exploration and exploitation based on the predictive posterior of the surrogate. 
Given predictive mean $\mu(x)$, standard deviation $\sigma(x)$, and the best 
observed value $f_{\best}$, common acquisition criteria include the {\bf lower confidence bound} ({\tt LCB}), {\bf probability of improvement} ({\tt PI}), and {\bf expected improvement} ({\tt EI}). To retain the minimization convention used throughout this paper, we define
\[
m(x,\theta) := \begin{cases}
    \mu(x)-\kappa \sigma(x), & \mbox{{\tt LCB},}\\
    -\Phi(z), & \mbox{{\tt PI},}\\
    -\big[(f_{\best}-\mu(x)-\xi)\Phi(z)+\sigma(x)\phi(z)\big], & \mbox{{\tt EI},}
\end{cases}
\]
where $z=(f_{\best}-\mu(x)-\xi)/\sigma(x)$ for $\sigma(x)>0$, and $\Phi(\cdot)$ and $\phi(\cdot)$ denote the CDF and PDF of the standard normal distribution, respectively. Zero-variance cases use the deterministic improvement criterion (with zero probability for a strict tie). The nonnegative parameters $\kappa$ and $\xi$ control the trade-off between exploration ($\sigma(x)$) and exploitation ($\mu(x)$). Optimizing $m(x,\theta)$ is itself a nontrivial subproblem, usually approached by multi-start local search or evolutionary heuristics.

{\tt GPyOpt} \cite{GPyOpt}, {\tt BOHB} \cite{BOHB}, {\tt Scikit-Optimize} \cite{Scikit-Optimize}, {\tt Spearmint} \cite{Snoek2012}, and {\tt Dragonfly} \cite{Dragonfly} are five representative {\tt BO} solvers. {\tt GPyOpt} includes a {\tt GP}-based {\tt BO} library, {\tt Spearmint} is a popular {\tt BO} framework for hyperparameter tuning, {\tt Scikit-Optimize}  includes a lightweight library for {\tt BO} with scikit-learn, {\tt BOHB} combines {\tt BO} with resource-aware HyperBand for scalable optimization, {\tt Dragonfly} is a scalable {\tt BO}  package for multi-fidelity and high-dimensional problems.

\subsection{Polling-Based Methods}\label{sec:DS}

The generic polling and line-search templates in the supplement describe unconstrained objective-decrease mechanisms. Applying them to~\eqref{e.CNLP} requires explicit feasibility or constraint handling.

In this section, we discuss two classical polling-based algorithms: \textbf{Direct Search ({\tt DS})} and \textbf{Line Search ({\tt LS-dd})} without approximate directional derivatives. Both methods rely purely on function evaluations along chosen directions and use a forcing function as a sufficient decrease condition, rather than constructing approximate gradients or directional derivatives. In this way, they explore the search space by polling trial points and accepting steps only if the forcing condition is satisfied. These methods need only function evaluations; noise robustness and convergence for discontinuous objectives require additional assumptions or dedicated safeguards.

The difference between these two algorithms is that {\tt LS-dd} performs an extrapolation step along a fixed direction as long as reductions in the objective function values are found while step sizes are increased.  The goal of extrapolation is to speed up the algorithm to find an approximate stationary point. 

Under appropriate smoothness and search-direction assumptions,
these methods admit first-order stationarity guarantees.
Unconstrained smooth problems can use a gradient-norm criterion;
constrained problems require an appropriate feasible or projected
stationarity measure. Such guarantees do not imply convergence
to local or global minimizers. Practical global-search performance
of particular variants is problem-dependent; see \cite{VRBBO}.

{\bf Direct Search ({\tt DS}) methods.} Algorithm 2 in {\tt SuppMat.pdf} is a generic version of the {\tt DS} algorithm.  {\tt DS} methods seek sufficient decrease by polling trial points along a prescribed set of directions. The search directions can be coordinate directions or random directions, or even any approximate descent directions if the gradient is approximated by fitting or a finite difference method. Three main versions of direct search methods are pattern search, the Nelder-Mead method, and mesh adaptive direct search. The state-of-the-art direct search solvers are {\tt NOMAD} \cite{NOMAD}, {\tt BFO} \cite{BFO}, {\tt NMSMAX} \cite{NMSMAX}, {\tt PSM} \cite{PSM}, and {\tt DSPFD} \cite{DSPFD}. {\tt NOMAD} is a mesh-adaptive direct search, which can handle small- to large-scale constrained mixed-integer DFO problems, but for large-scale problems is quite slow. {\tt BFO} uses a direct search algorithm, which handles small- to large-scale constrained mixed-integer DFO problems. {\tt NMSMAX} implements the Nelder--Mead simplex method, which can handle small- and medium-scale DFO problems. {\tt PSM} uses a pattern search algorithm, which is effective for small-scale DFO problems. {\tt DSPFD} uses randomized direct search for bound- and linearly constrained DFO problems. Their performance and noise-handling capabilities depend on the implementation and problem class.

{\bf Line Search ({\tt LS-dd}) methods.} {\tt LS-dd} uses no approximate directional derivative. Algorithm 3 in {\tt SuppMat.pdf} introduces a generic {\tt LS-dd} framework without approximate directional derivative or approximate gradient vector for the unconstrained objective-decrease setting.  {\tt LS-dd} uses the random or coordinate search directions. For the second class, like {\tt LS-dd}, the term $\alpha\nabla f(x_{\best})^Td$ is replaced by $-\alpha^2$ in the Armijo condition, resulting in the {\tt LS} condition $\wt \mu(\alpha)\ge c_1$ without the approximate directional derivative $\nabla  f(x_{\best})^T d$, where $\wt\mu(\alpha):=(f(x_{\best}+\alpha d) - f(x_{\best}))/(-\alpha^2)$. This {\tt LS} method uses an extrapolation step, which differs from the extrapolation step used in the line searches of the first class. In an extrapolation step in the second class, the extrapolation step sizes are expanded by a factor (larger than one), and the corresponding trial points and their function values are computed until the {\tt LS-dd} condition is violated. Then, a trial point with the smallest function value is chosen as the best point by such an extrapolation step.  Extrapolation seeks additional sufficient decrease along the selected direction.

\subsection{Local-Approximation-Based Methods}\label{sec:LS+dd}

In this section, we discuss two classes of local-approximation-based methods that exploit approximate first-order information: 
\textbf{Line Search with directional derivatives ({\tt LS+dd})} and \textbf{Trust-Region ({\tt TR})} algorithms. Unlike polling-based approaches, which rely solely on function values and forcing functions, these methods construct local models by approximating gradients or directional derivatives, either through finite differences or interpolation. The resulting local approximations are then used to determine descent directions and step sizes, supporting stationarity guarantees under suitable model-accuracy and regularity assumptions. At the same time, when combined with curvature information or regularization strategies, they can be extended to approximate local minimizers in nonlinear BBO problems.

Algorithm 4 in {\tt SuppMat.pdf} is a generic {\tt LS+dd} framework with an approximate directional derivative or approximate gradient for the unconstrained objective-decrease setting. {\tt LS+dd} can compute the gradient by fitting or a finite difference method, and the search direction can be any approximate descent directions if the gradient is approximated like approximate steepest descent, approximate conjugate gradient, or approximate quasi-Newton directions, and random or coordinate directions if the directional derivative is approximated by a finite difference method otherwise. This algorithm includes either a backtracking or an interpolation step (or extrapolation step). In the first class, we first define the {\bf Goldstein quotient} by
\[
\mu (\alpha) := \frac{f(x_{\best}+\alpha d) - f(x_{\best})}{\alpha \nabla f(x_{\best})^Td}\ \  \mbox{for}\ \ \alpha > 0.
\]
Then, we discuss four {\tt LS+dd} methods. To satisfy the Armijo condition $\mu(\alpha)\ge c_1$, backtracking may be used. The Goldstein condition
\[
c_1\le \mu(\alpha)\le c_2, \qquad 0<c_1<c_2<1,
\]
is two-sided and may require either decreasing or increasing the trial step. The Wolfe line search includes the Armijo condition and the curvature condition defined by
\[
\nabla f(x_{\best} + \alpha d)^T d \geq c_3 \nabla f(x_{\best})^T d \ \ \mbox{with $0 < c_1 < c_3 < 1$.}
\]
Until the Wolfe conditions are satisfied, an interpolation step or extrapolation step is performed. The improved Goldstein {\tt LS+dd} proposed in \cite{CLS} satisfies the sufficient descent condition
\[
\mu(\alpha)|\mu(\alpha)-1|\ge\beta \ \ \mbox{for some fixed $\beta\in]0,1/4[$.}
\]
This can be done in two stages: as long as the sufficient descent condition is violated, an interpolation step or an extrapolation step is performed to create an interval, and then the geometric mean of the lower and upper bounds of the interval is taken as the new step size.

{\tt TR} methods approximate nonlinear objective function by a linear, quadratic, or cubic model, which is restricted to a trust region to avoid large steps and increase the accuracy of the model. If the agreement between the objective function and the model function is good, the trial point is accepted as the new point and the trust region remains unchanged or expanded; otherwise, the trial point is discarded and the trust region is reduced. Representative model-based DFO solvers include {\tt BOBYQA} \cite{BOBYQA}, {\tt BCDFO} \cite{BCDFO}, {\tt UOBYQA} \cite{UOBYQA}, {\tt MATRS} \cite{MATRS}, and {\tt SNOBFIT} \cite{HuyN}. Among these, {\tt BOBYQA}, {\tt BCDFO}, and {\tt UOBYQA} are continuous trust-region methods, whereas {\tt MATRS} addresses noisy bound-constrained mixed-integer optimization using randomized model construction. {\tt SNOBFIT} is a model-based method designed for noisy black-box optimization.

Algorithm 1 in {\tt SuppMat.pdf} can be used as a structural template for a {\tt TR} framework, with its generic acceptance step replaced by the trust-region ratio-based update described below, for the unconstrained objective-decrease setting. At first, this algorithm generates well-distributed sample points using a space-filling method and evaluates their function values.  Then, it alternately constructs a local surrogate model $m(x,\theta)$, where the trust region is defined as  
\[
\mathcal R := \{x \in C_{\cont} \mid \|x - x_{\best}\| \le \Delta\}
\]  
with $C_{\cont}$ from \eqref{e.contSet} and the trust-region radius $\Delta$
(which initially is a positive tuning parameter).  The surrogate is first solved within $\mathcal R$ to obtain $x_{\trial}$, after which $f(x_{\trial})$ is evaluated. Here, $\theta$ encodes the model coefficients, while $\Delta$ determines the trust region.  The {\bf trust-region ratio}  
\[
\rho := \frac{f_{\best}-f_{\trial}}{m(x_{\best},\theta)-m(x_{\trial},\theta)}
\]  
measures agreement when the predicted reduction in its denominator is positive; otherwise reject the trial or improve the model.  
If $\rho \ge \eta \in (0,1]$, the step is considered successful, and $\Delta$ is 
unchanged or expanded as $\Delta \gets \lambda \Delta$ with $\lambda>1$. In this case, the best point and its function value are updated as $x_{\best}=x_{\trial}$ and $f_{\best}=f_{\trial}$. Otherwise, the step is unsuccessful, and $\Delta$ is reduced as $\Delta \gets \Delta/\lambda$.

\subsection{BBO and Its Applications}\label{sec:GWBBO}

In this section, we illustrate how BBO methods are applied across operations research, engineering, and ML. Our goal is to highlight both classical DFO use cases in OR and the generic BBO workflow that underpins modern applications, thereby motivating the role of BBO as a unifying paradigm for complex real-world problems.

{\bf DFO Applications in OR.} DFO methods are often used in OR to solve complex optimization problems where derivatives of the objective function or constraints are not available, expensive to calculate, or do not exist at all. These methods are particularly suitable for BBO problems where the objective function is evaluated by simulations, experiments, or computationally intensive models. 

In robust and stochastic optimization, uncertainties in {\bf decision-making models} can be addressed by surrogate-based DFO algorithms such as {\tt BO} (cf.~\cite{Garnett2023BO}). 
Another interesting application in {\bf energy systems optimization} is energy distribution networks, where black-box energy models with constraints (e.g., demand--supply balance and renewable integration) must be optimized using stochastic DFO techniques to handle uncertain energy outputs. 
DFO is also a powerful tool to capture the nonlinear, discrete, and uncertain properties of {\bf supply chain problems} (cf.~\cite{griffis2012metaheuristics,HubbsOR-Gym}), where brute-force enumeration is infeasible. 

Under the taxonomy of polling-based, surrogate-based, and local-approximation-based methods, different strategies are used in OR applications: polling-based methods (e.g., direct search, randomized line search) can be applied to simulation models, with noise robustness depending on the algorithm; surrogate-based methods (e.g., {\tt GP}s, {\tt BO}) are effective for costly simulation-driven planning; and local-approximation-based methods (e.g., trust-region models, gradient approximations) are particularly suited to small- and medium-scale models with relatively smooth structure.  

Beyond supply chains, DFO methods and their metaheuristic extensions are applied in {\bf logistics and transportation} \cite{griffis2012metaheuristics}, {\bf financial portfolio optimization} \cite{erwin2023meta}, {\bf manufacturing and production scheduling} \cite{jarboui2013metaheuristics}, and many other complex OR settings where explicit derivatives are unavailable or unreliable.

{\bf A Generic BBO Workflow.} Kumagai and Yasuda~\cite{kumagai2023bbo} survey the landscape of BBO and its practical applications in Artificial Intelligence (AI) and industry.  
BBO refers to the optimization of an objective $f(x)$ that is expensive to evaluate and provides 
no gradient or structural information.  
Typical applications include hyperparameter tuning, simulation-based design, robotics control, 
real-time decision-making, and large-scale industrial systems (e.g., energy, medical, and manufacturing).  

A generic BBO workflow typically begins by defining the objective function $f(x)$ and the feasible domain $\Omega$ (cf. Algorithm 5 in {\tt SuppMat.pdf}). The procedure starts with an initialization phase in which exploratory samples are generated, often using random or space-filling designs. An optimizer such as Bayesian optimization, evolutionary strategies, or a DFO method then proposes candidate solutions $x_{\trial}$. These candidates are evaluated through $f(x_{\trial})$, and the incumbent best solution is updated. When relevant, surrogate models or partition structures are refined to guide the search more effectively.

This general framework is widely applicable because it can handle objectives that are non-differentiable, noisy, or based on computationally expensive simulations. Surrogates, AI-driven models, and meta-learning approaches enable efficient reuse of past evaluations. Furthermore, the modular nature of the framework allows different optimizers to be plugged in easily and enables integration with HPC or simulation infrastructures, making it suitable for large-scale industrial settings. 

Thus, BBO serves as a unifying paradigm that bridges optimization theory with real-world 
deployment in AI and engineering.

\subsection{Recommendation and Conclusion}

We classified DFO solvers into three main categories: 
{\bf Polling-Based Methods}, {\bf Surrogate-Based Methods}, and {\bf Local-Approximation-Based Methods}. Table~\ref{t.DFO} summarizes representative algorithms in each category, 
together with qualitative indications of problem-size applicability and parallel execution support in suitable implementations. 

Polling-based solvers (e.g., {\tt NOMAD}, {\tt BFO}) are robust and derivative-free, 
but can become slow on medium- and large-scale problems unless parallelized 
(e.g., via {\tt MPI} \cite{MPI} or {\tt CUDA} \cite{CUDA}). 
Surrogate-based methods can reduce expensive evaluations, but
their scalability depends on the model, sampling strategy,
problem structure, and budget. Parallel evaluation can reduce
elapsed time while high-dimensional modeling remains challenging.
Local-approximation-based methods (e.g., {\tt SSDFO},
{\tt FMINUNC}, {\tt BOBYQA}) use gradient or model
approximations. Their performance depends on smoothness,
noise, dimension, and the available subspace structure.
Hybrid solvers can combine model-free and model-based steps.

\begin{table}[!htpp]
\begin{center}
\scalebox{0.85}{\begin{tabular}{|l|lllll|}
\hline
\multicolumn{1}{|l|}{{\bf category}} & \multicolumn{3}{c}{{\bf problem size}}  & \multicolumn{1}{l}{{\bf HPC?}} &  \multicolumn{1}{l|}{{\bf software/references}} \\
 &  \rot{{\bf small}} & \rot{{\bf medium}} & \rot{{\bf large}} & & \\
\hline 
{\bf Polling-Based} &  $+$ & $+$ & $+$ & $\pm$ & 
\begin{tabular}{l}
{\tt NOMAD} \cite{NOMAD}, {\tt BFO} \cite{BFO}, \\
{\tt NMSMAX} \cite{NMSMAX}, {\tt PSM} \cite{PSM}, \\
{\tt DSPFD} \cite{DSPFD}, {\tt DFLINT} \cite{DFLINT}, \\
{\tt DFLBOX} \cite{DFLBOX}, {\tt DFNDFL} \cite{DFNDFL},\\
{\tt VRBBO} \cite{VRBBO}, {\tt SDBOX}  \cite{SDBOX}, \\
{\tt VRDFON} \cite{VRDFON}
\end{tabular}\\
\hline 
{\bf Surrogate-Based} & $+$ & $+$ & $\pm$ & $\pm$& 
\begin{tabular}{l}
{\tt pySOT} \cite{pySOT}, {\tt Dakota} \cite{Dakota}, \\
{\tt SPLINE} \cite{SPLINE}, {\tt MISO} \cite{Mller2015}, \\
{\tt GPyOpt} \cite{GPyOpt}, {\tt BOHB} \cite{BOHB},\\
{\tt Scikit-Optimize} \cite{Scikit-Optimize}, \\
{\tt Spearmint} \cite{Snoek2012}, \\
{\tt Dragonfly} \cite{Dragonfly}, \\
{\tt SNOBFIT} \cite{HuyN}
\end{tabular}\\
\hline 
{\bf Local-Approximation-Based} &  $+$ & $+$ & $\pm$ &$\pm$& 
\begin{tabular}{l}
{\tt SSDFO} \cite{SSDFO}, {\tt FMINUNC} \cite{FMINUNC}, \\
{\tt BOBYQA} \cite{BOBYQA}, {\tt BCDFO} \cite{BCDFO},\\
{\tt UOBYQA} \cite{UOBYQA}, {\tt MATRS} \cite{MATRS}\\
\end{tabular}\\
\hline
\end{tabular}}
\end{center}
\caption{Qualitative, overlapping classification of DFO
solvers. A $+$ indicates applicability in suitable formulations;
$\pm$ indicates stronger problem or implementation dependence.
Size categories are relative, not universal dimension cutoffs.
Parallel support depends on the implementation; this table does
not report a common performance benchmark.}
\label{t.DFO}
\end{table}

\section{Learning-Based Enhancements for BBO}\label{sec:NNs-BBO}

Machine learning (ML) and reinforcement learning (RL) have emerged as powerful tools for enhancing BBO, where the objective and constraints can only be accessed through expensive function evaluations, and gradient information is unavailable.  Classical DFO methods often struggle in such settings due to the combinatorial growth of discrete assignments, the complexity of continuous domains, and the high cost of evaluations.  

As expressive function approximators, adaptive optimizers, meta-learners, and generative models, learning-based methods introduce new mechanisms for guiding the search process, 
reusing data efficiently, and scaling optimization to complex domains.  In the following, we survey representative learning-driven approaches that enhance BBO, 
including surrogate-based formulations solved via modular model-based frameworks~\cite{bischl2018mlrmbo}, 
adaptive momentum methods~\cite{chen2019zo}, 
benchmarking-informed portfolios and distributed search frameworks~\cite{meunier2022abbo,cuccu2022dibb}, 
and neural generative optimizers~\cite{li2025b2opt, li2024diffusionbbo}. Classical {\tt DFO-TR}, optimizer-inspired {\tt ZO-AdaMM}, and distributed {\tt DiBB} are included as related frameworks; they do not themselves require a learned neural search policy.

The methods described in Section~\ref{sec:BBO} 
(line search, direct search, and model-based solvers, with {\tt BO} as a special case of the latter) constitute the {\bf classical foundation} of BBO. They provide general strategies for searching without gradients, but each faces limitations in practice: line search and direct search may scale poorly in high dimensions; trust-region surrogates may struggle with non-smooth or categorical domains; and {\tt BO} is limited by surrogate expressiveness and acquisition optimization.

\medskip
ML and RL do not introduce entirely new classes of BBO solvers, but instead 
{\bf enhance existing ones} by providing richer models, adaptive updates, and data-driven strategies.  
For example:
\begin{itemize}
  \item ML-based surrogates (e.g., 
        \texttt{mlrMBO}~\cite{bischl2018mlrmbo}) extend the surrogate modeling paradigm of {\tt BO} and model-based DFO.  
  \item Optimizer-inspired updates (e.g., {\tt ZO-AdaMM}~\cite{chen2019zo}) import adaptive learning-rate and momentum techniques from ML into gradient-free search,  improving robustness over classical line/direct search.  
  \item Benchmarking-informed portfolios such as {\tt ABBO}
  \cite{meunier2022abbo} select or chain optimizers. Learned
  partitioning ({\tt SPBOpt} \cite{sazanovich2021spbopt}) and
  distributed decomposition ({\tt DiBB} \cite{cuccu2022dibb})
  organize search through different mechanisms.  
  \item Generative ML models (e.g., {\tt B2Opt}~\cite{li2025b2opt}, 
        {\tt DiffBBO}~\cite{li2024diffusionbbo}) 
        offer new ways of sampling candidate solutions, 
        complementing traditional acquisition optimization in surrogate-based methods.  
  \item At the RL--BBO interface, {\tt RBO}~\cite{choromanski2019rbo},
        {\tt CAS-MORE}~\cite{huettenrauch2024more}, and
        {\tt LB-SGD}~\cite{usmanova2024logsafe} address policy search
        and constrained optimization, while {\tt Surr-RLDE}~\cite{ma2025surr-rlde}
        and {\tt Q-Mamba}~\cite{ma2025qmamba} learn policies for configuring search.
\end{itemize}

\medskip
Thus, the ML and RL sections can be viewed as 
a {\bf second layer} built on top of the classical BBO taxonomy: classical BBO defines the optimization backbone, 
while ML and RL provide modern enhancements 
to make these solvers more scalable, robust, and adaptive.

\subsection{ML Enhancements in BBO}\label{sec:MLenhance}

ML provides a versatile set of tools to enhance 
BBO, where search must be conducted without explicit gradients and often under the complexity of combinatorial, continuous, or mixed domains. While classical approaches, such as {\tt BO} and {\tt ES}, offer general-purpose strategies, ML introduces richer models and adaptive mechanisms that improve efficiency, scalability, and robustness.  

Broadly, ML contributes to BBO in four complementary ways:  
(i) through {\bf surrogate modeling}, where expressive regressors approximate the expensive objective; (ii) through {\bf optimizer-inspired updates}, which transfer techniques such as adaptive momentum from deep learning to the zeroth-order setting;  
(iii) through {\bf meta-learning and algorithm portfolios}, which select or adapt optimizers across tasks; and  (iv) through {\bf generative neural models}, which directly learn distributions over promising solutions.

In the following subsections, we review eight representative methods that illustrate these roles, ranging from  \texttt{mlrMBO} to recent advances such as {\tt B2Opt} and {\tt DiffBBO} (for more details see Table~\ref{t.MLBBO}). 

\begin{table}[!ht]
\begin{center}
\scalebox{0.86}{
\begin{tabular}{|l|l|l|l|}
\hline
{\bf Method} & {\bf Core Idea} & {\bf Domain} & {\bf Refs.} \\
\hline
{\tt mlrMBO} & Modular {\tt SMBO} with flexible surrogates & Mixed/Continuous & \cite{bischl2018mlrmbo} \\
{\tt ZO-AdaMM} & Zeroth-order Adam with adaptive moments & Continuous/Noisy & \cite{chen2019zo} \\
{\tt ABBO} & Algorithm selection and chaining & General BBO & \cite{meunier2022abbo} \\
{\tt DiBB} & Distributed block-wise optimization & High-dimensional & \cite{cuccu2022dibb} \\
{\tt SPBOpt} & Partition-based local {\tt BO} for low budgets & Low-budget BBO & \cite{sazanovich2021spbopt} \\
{\tt DFO-TR} & Trust-region DFO applied to ML objectives & Continuous/Noisy & \cite{ghanbari2017dfotr} \\
{\tt B2Opt} & Transformer crossover/mutation/selection & Low-budget BBO & \cite{li2025b2opt} \\
{\tt DiffBBO} & Reward-conditioned diffusion sampling & Offline/Data-driven & \cite{li2024diffusionbbo} \\
\hline
\end{tabular}}
\end{center}
\caption{Representative ML-related and supporting BBO frameworks.}
\label{t.MLBBO}
\end{table}

In this section, we draw on standard components from BBO, ML, and RL.  We refer to surrogate models such as Gaussian Processes ({\tt GP}s), Random Forests, and Gradient Boosting ({\tt GB}); population-based optimizers including Evolution Strategies ({\tt ES}), Covariance Matrix Adaptation Evolution Strategy ({\tt CMA\text{-}ES}), Particle Swarm Optimization ({\tt PSO}), and Differential Evolution ({\tt DE});  and acquisition functions from Bayesian optimization ({\tt BO}) such as Expected Improvement ({\tt EI}), Lower Confidence Bound ({\tt LCB}), and Probability of Improvement ({\tt PI}).  
We also make use of distributed formulations like Distributed Block-Wise Optimization ({\tt DiBB}) and neural modules appearing in generative optimizers—Self-Attention Crossover ({\tt SAC}), Feed-Forward Mutation ({\tt FM}), and the Residual and Selection Module ({\tt RSSM}) used in {\tt B2Opt}. Common performance metrics include the Receiver Operating Characteristic ({\tt ROC}) curve and its Area Under the Curve ({\tt AUC}).  
Zeroth-order optimization ({\tt ZO}) also appears in several methods, together with adaptive variants such as {\tt AdaMM}. The supplementary material ({\tt SuppMat.pdf}) provides full definitions, background explanations, and pseudocode for all ML-related components referenced in this section, allowing readers to consult a single source for precise technical details.

\subsubsection{\texttt{mlrMBO} -- Modular Model-Based Optimization}\label{sec:mlrMBO}

The Modular Model-Based Optimization (\texttt{mlrMBO}) framework~\cite{bischl2018mlrmbo} 
is a general-purpose implementation of surrogate-based optimization, also known as 
Sequential Model-Based Optimization ({\tt SMBO}).  {\tt BO} is a prominent special case within this broader {\tt SMBO} paradigm.  
\texttt{mlrMBO} was designed for real-world settings with mixed-variable domains 
(continuous, integer, categorical, and conditional parameters), multi-objective tasks, and parallel batch evaluations. The pseudocode of the \texttt{mlrMBO} is Algorithm 6 in Subsection 2.1.2 of {\tt SuppMat.pdf}.

The ML contribution of \texttt{mlrMBO} lies in its highly flexible surrogate modeling: any regression learner from the modular ML toolbox can serve as the surrogate model, including Random Forests, \texttt{GP}, \texttt{GB}, and others. Random Forests are often used in heterogeneous domains with continuous and categorical variables, as they naturally handle non-continuous inputs without requiring explicit embeddings. \texttt{GP}s with custom kernels, such as \texttt{RBF}, categorical, and conditional kernels, are supported for continuous problems. Moreover, the modular design allows users to replace the surrogate model, the acquisition function, or the optimizer as needed, depending on the characteristics of the problem.

The flexibility of its surrogate models makes \texttt{mlrMBO} effective across a wide range of BBO tasks, from engineering simulations to hyperparameter tuning and algorithm configuration. Its main features include support for continuous, integer, categorical, and hierarchical (conditional) parameters; the ability to propose multiple candidate points per iteration for efficient parallelization; both dominance-based and scalarization-based acquisition strategies; built-in handling of noisy evaluations, failed runs, and missing data; and tools for monitoring optimization progress through integrated ML capabilities. 

\texttt{mlrMBO} has been successfully applied to hyperparameter optimization in ML pipelines ({\tt SVMs}, {\tt NNs}, ensembles),  multi-objective tuning of runtime–accuracy trade-offs in solvers,   algorithm configuration for combinatorial optimization problems, and engineering design with expensive simulation-based evaluations.

\subsubsection{{\tt ZO-AdaMM} -- Zeroth-Order Adaptive Momentum Method}\label{sec:ZO-AdaMM}

A prominent example of how ideas from \textbf{ML optimizers} can enhance BBO 
is the Zeroth-Order Adaptive Momentum Method ({\bf {\tt ZO-AdaMM}}) by Chen et al.~\cite{chen2019zo}. 
{\tt ZO-AdaMM} transfers {\tt Adam}/{\tt AMSGrad}-style {\bf adaptive moment estimation} to the gradient-free setting by
replacing analytic gradients with randomized zeroth-order ({\tt ZO}) estimates and using an \texttt{AMSGrad} elementwise running maximum of second moments
together with a {\bf Mahalanobis} projection for constraints. The pseudocode of \texttt{ZO-AdaMM} is Algorithm 7 in Subsection 2.1.3 of {\tt SuppMat.pdf}.

This is motivated by black-box ML tasks (adversarial example generation, RL, hyperparameter tuning)
where gradients are unavailable or unreliable. Classical {\tt ZO} methods (e.g., {\tt ZO-SGD}, coordinate descent) use fixed steps
and suffer high variance in high dimensions. By importing adaptive moments $(m_t,v_t)$, {\tt ZO-AdaMM} improves robustness and stability.

Key improvements in \texttt{ZO-AdaMM} include the use of random-direction gradient surrogates constructed from function evaluations at perturbed inputs, eliminating the need for analytic gradients. The method employs \texttt{Adam}/\texttt{AMSGrad}-style moving averages $(m_t, v_t)$, where an elementwise running maximum of second moments supplies nondecreasing adaptive scaling. For constrained problems, it relies on Mahalanobis-distance projections, since standard Euclidean projections may fail to converge. Theoretical results show convergence rates whose dependence on the dimension is roughly $O(\sqrt{d})$ worse than that of first-order adaptive methods, which is consistent with expectations in zeroth-order settings.

Empirically (e.g., on ImageNet attacks, both per-image and universal), {\tt ZO-AdaMM} converges faster and attains higher success rates than {\tt ZO} baselines ({\tt ZO-SGD}, {\tt ZO-signSGD}, {\tt ZO-SCD}), exhibiting the practical value of bringing deep-learning optimizer design into gradient-free ML.

\subsubsection{{\tt ABBO} -- Algorithm Selection Wizard for BBO}\label{sec:ABBO}

BBO includes continuous, mixed-variable, noisy, and multi-objective
tasks. Performance varies with the problem class, evaluation
budget, and noise characteristics, motivating algorithm selection
and portfolios.  

To address this, Meunier et al.~\cite{meunier2022abbo} propose the {\bf Automated Black-Box Optimizer ({\tt ABBO})}, 
implemented in the Nevergrad platform (also called {\tt NGOpt}). {\tt ABBO} is a meta-algorithm that leverages massive benchmarking data to select, 
combine, or adapt optimizers during a run. The pseudocode of \texttt{ABBO} is Algorithm 8 in Subsection 2.1.4 of {\tt SuppMat.pdf}.  

\texttt{ABBO} leverages data-driven insights from large-scale benchmarking to guide both algorithm selection and parameter control. It performs passive algorithm selection by using simple problem meta-features such as dimension, variable types, noise level, degree of parallelism, and evaluation budget to choose optimizers that have historically performed well on similar tasks. It also supports an active testing phase, similar to a bet-and-run strategy, in which several algorithms are briefly tried on the target problem before continuing with the most promising one. In addition, \texttt{ABBO} can chain multiple optimizers within a single run, for example, beginning with random search and later switching to \texttt{CMA-ES} for refinement, following rules derived from extensive benchmarking studies.

{\tt ABBO} therefore uses selection and chaining heuristics
informed by extensive benchmarking.

\subsubsection{{\tt DFO-TR} -- BBO in ML with Trust-Region DFO}\label{sec:DFO-TR}

Ghanbari and Scheinberg~\cite{ghanbari2017dfotr} study the application of the 
\textbf{Derivative-Free Optimization with Trust Regions ({\tt DFO-TR})} algorithm 
to  ML tasks where explicit gradients are unavailable or unreliable.  
Their motivating examples include optimizing non-smooth objectives such as {\tt AUC} and hyperparameter tuning of ML models.  The pseudocode of a generic version of \texttt{DFO-TR} is Algorithm 9 in Subsection 2.1.5 of {\tt SuppMat.pdf}. 

\texttt{DFO-TR} is a model-based trust-region method. At each iteration, it constructs a local surrogate model $m(x)$ inside a trust region centered at the current iterate. The cited algorithm constructs a quadratic interpolation model from evaluated points and maintains an interpolation set as the search proceeds. The surrogate is then optimized, approximately, to produce a trial point $x_{\trial}$, which is evaluated on the true black-box function $f(x)$. The trust-region radius is then updated based on the agreement between the surrogate’s predicted improvement and the actual improvement.

This approach is especially effective for ML optimization tasks because it can handle noisy, non-differentiable objectives such as \texttt{AUC} or validation error, where gradients are unavailable. It also makes efficient use of past evaluations, unlike purely random or grid searches, and it usually requires far fewer function evaluations, which is important when training ML models is computationally expensive.

Empirical results show that {\tt DFO-TR} can efficiently maximize {\tt AUC} in classification tasks (e.g., linear classifiers, {\tt SVMs}) without requiring gradients, and can tune hyperparameters (regularization, kernel width) with performance competitive to {\tt BO}, random search, and gradient-based heuristics.

\subsubsection{{\tt SPBOpt} -- Solving BBO via Learning Search Space Partition}\label{sec:SPBOpt}

Sazanovich et al.~\cite{sazanovich2021spbopt} propose the 
\textbf{Search Partition for Bayesian Optimization ({\tt SPBOpt})} algorithm, developed during the NeurIPS 2020 BBO Challenge.  {\tt SPBOpt} addresses the challenge of {\bf low-budget optimization}, where only a small number of 
function evaluations is allowed.  
The key idea is to learn a partition of the search space and apply local {\tt BO} within a selected promising region, rebuilding the partition as evaluations proceed. The pseudocode of a generic version of \texttt{SPBOpt} is Algorithm 10 in Subsection 2.1.6 of {\tt SuppMat.pdf}.    

After initial sampling, \texttt{SPBOpt} learns a partition from
the observations and selects the region with the lowest average
observed objective value. Modified local \texttt{TuRBO} then
proposes candidates there. The partition is periodically rebuilt,
with a restart after prescribed stagnation.

This approach works especially well in competition and benchmark settings because local \texttt{BO} is highly sample-efficient, which is essential when only a small evaluation budget is available. The use of partitions helps balance exploration and exploitation by spreading queries across different regions, and the adaptive refinement of these partitions allows the method to concentrate on the most promising subspaces as more evidence is gathered. {\tt SPBOpt} ranked third in the NeurIPS 2020 BBO Challenge finals, demonstrating strong empirical performance 
under strict evaluation budgets.

Empirical results reported by Sazanovich et al.~\cite{sazanovich2021spbopt} indicate that \texttt{SPBOpt} performed strongly, ranking third overall in the NeurIPS 2020 BBO Challenge finals and surpassing many competitive baselines under the strict 128-evaluation budget. The partitioning combined with local \texttt{BO} proved particularly effective on low-dimensional and highly multi-modal problems, where broad global coverage together with adaptive refinement enabled the method to quickly concentrate on promising regions. When compared with standard \texttt{BO}, \texttt{SPBOpt} delivered higher-quality solutions within the same limited budget by maintaining exploration across partitions while exploiting locally within the selected subregion.

\subsubsection{{\tt B2Opt} -- Learning to Optimize BBO with Little Budget}\label{sec:B2Opt}

Li et al.~\cite{li2025b2opt} propose \textbf{{\tt B2Opt}}, a deep learning framework for BBO under extremely limited evaluation budgets.  {\tt B2Opt} builds on ideas from genetic algorithms and Transformers, learning parameterized optimization strategies that map random initial populations to near-optimal solutions with few queries.  The pseudocode of a generic version of \texttt{B2Opt} is Algorithm 11 in Subsection 2.1.7 of {\tt SuppMat.pdf}.   

The main workflow begins by encoding the candidate population with a transformer-based network. Each \textbf{\texttt{B2Opt} Block (OB)} applies three components in sequence: a Self-Attention-based Crossover module (\texttt{SAC}), a Feed-Forward Mutation module (\texttt{FM}), and a Residual and Selection module (\texttt{RSSM}). Multiple OBs are stacked to update the population over successive iterations. The \texttt{B2Opt} model is trained on inexpensive surrogate problems so that it can approximate the behavior of the target BBO tasks, and the resulting learned optimizer is then applied to expensive black-box functions.

\texttt{B2Opt} delivers state-of-the-art performance in very low-budget optimization settings. On the standard \texttt{BBOB} benchmarks, it surpasses strong baselines such as \texttt{DE}, \texttt{CMA-ES}, and various \texttt{BO}-based methods. In 10-dimensional problems, it ranks first on 20 of the 24 \texttt{BBOB} functions, and in 100 dimensions it leads on 19 of the 24, demonstrating strong scalability. When compared with other learning-to-optimize approaches such as \texttt{LGA}, \texttt{LES}, and \texttt{L2O-swarm}, \texttt{B2Opt} consistently achieves better solution quality under the same evaluation budget. On real-world tasks, including robotics control and neural network training, it can discover high-quality solutions under the limited evaluation budgets reported for the corresponding experiments.

\subsubsection{{\tt DiffBBO} -- Diffusion Model for Data-Driven BBO}\label{sec:DiffBBO} 

Li et al.~\cite{li2024diffusionbbo} propose Reward-Conditioned Generation via Diffusion Models ({\tt RCGDM}), denoted by {\tt DiffBBO} in this survey, which casts offline BBO as conditional sampling: learn $p(x\mid r)$ from mixed labeled/unlabeled data via reward (or preference) modeling and pseudo-labeling; then generate near-optimal $x$ by conditioning on high reward. Under their statistical and latent-structure assumptions, they derive suboptimality bounds with a near-optimal off-policy-bandit reward-estimation term and analyze latent-subspace fidelity. The pseudocode of a generic version of \texttt{DiffBBO} is Algorithm 12 in Subsection 2.1.8 of {\tt SuppMat.pdf}.

The reported experiments study synthetic product design with
ambient dimension $64$ and latent dimension $16$, and directed
text-to-image generation. They examine reward improvement,
distribution shift, and latent-structure preservation
\cite{li2024diffusionbbo}.

\subsubsection{{\tt DiBB} -- Distributed Partially-Separable BBO}\label{sec:DiBB}

Cuccu et al. \cite{cuccu2022dibb} present {\bf {\tt DiBB}}, which turns a base BBO into a {\bf partially separable} distributed algorithm by partitioning parameters into blocks (e.g., by neural layers) and running independent solver instances per block, while assembling/evaluating full candidates centrally. It retains base-optimizer features such as step-size adaptation and can reduce elapsed time through distributed computation. The pseudocode of a generic version of \texttt{DiBB} is Algorithm 13 in Section 2.1.9 of \texttt{SuppMat.pdf}.

Empirical results reported by Cuccu et al.~\cite{cuccu2022dibb} show that \texttt{DiBB} can deliver substantial practical gains. On the \texttt{COCO}/\texttt{BBOB} benchmarks, it achieves up to a fivefold reduction in wall-clock time compared to the underlying solvers while maintaining their performance characteristics. In neuroevolution tasks, such as \texttt{Walker2D} with more than 11,000 parameters, \texttt{DiBB} scales effectively by distributing parameters across neural layers, allowing efficient optimization of extremely high-dimensional controllers. The reported examples demonstrate useful speedups; performance depends on the decomposition, coupling, base optimizer, and computing resources.

\subsection{RL and BBO}\label{sec:RLenhance}

RL and BBO interact in two directions. In RL-based optimizer
control, a learned policy configures search, as in {\tt Surr-RLDE}
and {\tt Q-Mamba}. In black-box policy search, BBO optimizes
policy parameters, as in {\tt RBO}, {\tt CAS-MORE}, and
{\tt PIBB}; {\tt LB-SGD} also addresses constrained policy
optimization. The following groups describe this broader interface:  
\begin{itemize}
  \item {\bf Robustness and safety}: robust regression, log-barrier formulations, 
        and KL-decoupled updates stabilize optimization under noise and constraints, 
        as in {\tt RBO}~\cite{choromanski2019rbo}, {\tt LB-SGD}~\cite{usmanova2024logsafe}, 
        and {\tt CAS-MORE}~\cite{huettenrauch2024more}.  
  \item {\bf Dynamic operator configuration}: RL policies adapt evolutionary 
        operators such as crossover and mutation during the search, 
        exemplified by {\tt Surr-RLDE}~\cite{ma2025surr-rlde}.  
  \item {\bf Policy search connections}: weighted policy-search
  updates reveal connections with evolution strategies, as
  illustrated by {\tt PI2}/{\tt PIBB}~\cite{stulp2012pibb}.  
  \item {\bf Meta-RL for algorithm configuration}: learned controllers generalize across heterogeneous tasks, 
        with offline training and benchmarks enabling zero-shot adaptation, as in {\tt Q-Mamba}~\cite{ma2025qmamba}, supported by {\tt MetaBox}~\cite{ma2023metabox} 
        and {\tt MetaBBO-RL}~\cite{ma2023metabox}.  
\end{itemize}

The following subsections review six representative methods that illustrate these roles: 
{\tt Surr-RLDE}, {\tt RBO}, {\tt CAS-MORE}, {\tt LB-SGD}, {\tt PI2}/{\tt PIBB}, and {\tt Q-Mamba} 
(cf. Table \ref{t.RLBBO}).

\begin{table}[!ht]
\begin{center}
\scalebox{0.815}{
\begin{tabular}{|l|l|l|l|}
\hline
{\bf Method} & {\bf Core Idea} & {\bf Domain} & {\bf Refs.} \\
\hline
{\tt RBO} & \begin{tabular}{l} 
Robust regression for \\
noisy gradient estimates 
\end{tabular}
& Noisy BBO/RL & \cite{choromanski2019rbo} \\
{\tt LB-SGD} & \begin{tabular}{l}
Log-barrier {\tt SGD} with\\
safe adaptive steps \end{tabular}
& Constrained BBO/RL & \cite{usmanova2024logsafe} \\
{\tt CAS-MORE} & 
\begin{tabular}{l} KL-decoupled mean/covariance \\
and entropy control 
\end{tabular} & Stochastic BBO/Episodic RL & \cite{huettenrauch2024more} \\
{\tt Surr-RLDE} & \begin{tabular}{l} RL configures {\tt DE} operators with \\
surrogate-based policy training 
\end{tabular} & {\tt MetaBBO} & \cite{ma2025surr-rlde} \\
{\tt PI2}/{\tt PIBB} & \begin{tabular}{l} Cost-weighted policy\\ updates \end{tabular}
& Evolutionary BBO/Policy search & \cite{stulp2012pibb} \\
{\tt Q-Mamba} & \begin{tabular}{l} Offline decomposed Q-learning \\
and Mamba backbone 
\end{tabular} & Large-action-space {\tt MetaBBO} & \cite{ma2025qmamba} \\
\hline
\end{tabular}}
\end{center}
\caption{Representative methods at the interface of RL and BBO.}
\label{t.RLBBO}
\end{table}

In this section, we draw on several key concepts from RL that are commonly incorporated into hybrid BBO methods.  
We refer to surrogate models such as the Kolmogorov--Arnold Network ({\tt KAN}) and training objectives like the Order-Aware Loss ({\tt OAL}); algorithmic families including Policy Optimization ({\tt PO}), Meta-Policy Online training, and Dynamic Algorithm Configuration ({\tt DAC}); and scalable formulations based on Decomposed Q-Functions ({\tt DQF}).  
Several RL-driven meta-optimizers also appear, such as Deep Evolutionary Double Q-Network ({\tt DEDQN}), Generalizable Learning of Evolutionary Operator Control ({\tt GLEET}), and Symbolic Operator Learning ({\tt SYMBOL}).  
For evaluation, we reference widely used benchmark suites and platforms, including {\tt COCO} (Comparing Continuous Optimizers), {\tt Pyomo} (Python Optimization Modeling Objects), {\tt Photonics} design tasks, {\tt LSGO} (Large-Scale Global Optimization), and the {\tt MuJoCo} simulator.  
The supplementary material ({\tt SuppMat.pdf}) provides full definitions, background explanations, and pseudocode for all RL-related components mentioned here, allowing readers to consult a single source for precise technical details.

\subsubsection{{\tt Surr-RLDE} -- RL-Configured {\tt DE} with Surrogate Training}\label{sec:Surr-RLDE} 

Ma et al.~\cite{ma2025surr-rlde} propose the \textbf{{\tt Surr-RLDE}} framework, 
which unifies surrogate learning with RL for Meta-Black-Box Optimization ({\tt MetaBBO}).  
The central idea is to train a {\tt KAN} surrogate with a relative-order-aware loss to preserve the ranking of candidate solutions.  
The paper evaluates {\tt KAN} surrogates for approximating its training problem instances.  This surrogate, trained per problem instance, replaces the most costly black-box evaluations during policy training, allowing an RL agent to learn how to dynamically configure 
the mutation operators and parameters in {\tt DE}.  The pseudocode of a generic version of \texttt{Surr-RLDE} is Algorithm 14 in Subsection 2.2.2 of {\tt SuppMat.pdf}. 

Fit one {\tt KAN} surrogate per training problem. Train a
Double-DQN controller on these fixed surrogates to select a
{\tt DE} mutation strategy and strength using progress rewards.
Deploy the frozen controller using true-objective evaluations.

This approach is effective because the RL component learns adaptive operator-selection policies that respond to the current search dynamics. It adapts mutation configurations during search, while surrogate environments reduce true-objective evaluations during policy training.

Extensive experiments show that {\tt Surr-RLDE} significantly reduces evaluation cost 
while maintaining competitive performance.  
It generalizes to higher-dimensional problems, 
matching or outperforming recent {\tt MetaBBO} baselines such as {\tt DEDQN}, {\tt GLEET}, and {\tt SYMBOL}~\cite{ma2025surr-rlde,ma2023metabox,song2024ribbo}.

\subsubsection{{\tt RBO} -- Provably Robust BBO for RL}\label{sec:RBO}  

Choromanski et al.~\cite{choromanski2019rbo} propose \textbf{Robust Black-box Optimization ({\tt RBO})}, a DFO method for RL that remains effective under adversarial or stochastic noise.  {\tt RBO} addresses the challenge of estimating reliable gradients when function evaluations are corrupted or unstable.  The pseudocode of a generic version of \texttt{RBO} is Algorithm 15 in Section 2.2.3 of \texttt{SuppMat.pdf}.

The main workflow of \texttt{RBO} starts by sampling a set of random perturbation directions around the current parameter vector. The reward function is evaluated at these perturbed points, even if some of the resulting measurements are noisy, unreliable, or partially corrupted. Based on these evaluations, \texttt{RBO} reconstructs an estimate of the local gradient by solving a robust regression problem that downweights outliers and adversarial noise. This gradient estimate is then used to update the policy parameters through an off-policy ascent step, allowing previously collected samples to be reused efficiently.

This approach is particularly effective because it can provably tolerate settings in which up to $23\%$ of the relevant measurements are arbitrarily corrupted, under the assumptions of the analyzed LP-decoding scheme. It also reuses past samples to produce stable gradient estimates, which helps reduce sample complexity. In addition, it delivers reliable optimization performance in noisy RL environments where evolution strategies and other derivative-free methods often fail.

Experiments on {\tt MuJoCo} robot control and quadruped locomotion confirm that {\tt RBO} continues to train policies successfully under heavy noise, where other blackbox methods fail.

\subsubsection{{\tt CAS-MORE} -- Coordinate-Ascent MORE with Step Size Adaptation}\label{sec:CAS-MORE} 

Hüttenrauch and Neumann~\cite{huettenrauch2024more} introduce 
\textbf{Coordinate-Ascent MORE with Step Size Adaptation ({\tt CAS-MORE})}, 
an improved version of the Model-based Relative Entropy Stochastic Search ({\tt MORE}) algorithm.  
{\tt CAS-MORE} extends {\tt MORE} to episodic RL with stochastic rewards, where ranking-based optimizers often fail.  The pseudocode of a generic version of \texttt{CAS-MORE} is Algorithm 16 in Subsection 2.2.4 of {\tt SuppMat.pdf}.

The main workflow of \texttt{CAS-MORE} begins by sampling candidate solutions from a Gaussian search distribution. A quadratic surrogate model of the objective is then fitted using ordinary least squares with appropriate preprocessing. The algorithm updates the mean and covariance separately, each constrained by its own KL-divergence bound, and dynamically adjusts the entropy through an evolution-path heuristic.

This design is effective because separating the updates of the mean and covariance improves both stability and convergence. The adaptive entropy schedule speeds up learning while preventing the exploration distribution from collapsing too early, and the streamlined surrogate-fitting procedure offers greater reliability when evaluations are noisy or stochastic.

{\tt CAS-MORE} outperforms ranking-based methods such as {\tt CMA-ES} and other policy-gradient-inspired black-box algorithms in episodic RL, achieving faster convergence and greater robustness to noise.

\subsubsection{{\tt LB-SGD} -- Log Barriers for Safe RL}\label{sec:LB-SGD} 

Usmanova et al.~\cite{usmanova2024logsafe} propose 
\textbf{Log-Barrier Stochastic Gradient Descent ({\tt LB-SGD})}, 
a method for smooth constrained stochastic optimization with safe-RL applications. Its zeroth-order variant uses only function feedback. Under the stated assumptions, safe initialization and the prescribed step-size and sampling rules keep iterates feasible with high probability. The pseudocode of a generic version of \texttt{LB-SGD} is Algorithm 17 in Subsection 2.2.5 of {\tt SuppMat.pdf}. 

The main workflow of \texttt{LB-SGD} starts by initializing the policy parameters and forming the log-barrier–augmented objective. A stochastic oracle is then queried, either through zeroth-order function values or noisy first-order gradients. The policy parameters are updated using stochastic gradient descent with a step size that adapts to the smoothness constant, and feasibility is controlled by the prescribed safe step-size and sampling rules.

The safety guarantee combines the barrier with safe initialization, oracle assumptions, and the prescribed adaptive step-size and sampling rules. Its adaptive step-size rules give provable convergence guarantees across convex, strongly convex, and even nonconvex settings. It also scales well to high-dimensional policy spaces, where safe Bayesian optimization becomes impractical, and it offers a lightweight approach to safe RL without requiring explicit constraint models.

{\tt LB-SGD} achieves state-of-the-art safe policy optimization, balancing exploration and safety in constrained RL, and outperforms prior safe {\tt BO} methods in high-dimensional tasks.

\subsubsection{{\tt PI2} vs {\tt PIBB} -- Policy Improvement between BBO and RL}\label{sec:PIBB}

Stulp and Sigaud~\cite{stulp2012pibb} investigate the connection between 
\textbf{Policy Improvement with Path Integrals ({\tt PI2})} 
and its simplified form \textbf{Policy Improvement with Black-Box ({\tt PIBB})}.  
They show that these methods are closely related to {\tt CMA-ES} and natural evolution strategies, revealing a bridge between policy search in RL and BBO.  The pseudocode of a generic version of \texttt{PIBB} is Algorithm 18 in Subsection 2.2.6 of {\tt SuppMat.pdf}.

The main workflow of \texttt{PIBB} starts by initializing the policy parameters.  The method then generates $K$ perturbed rollouts using parameters of the form policy parameters plus a small perturbation. After computing the cumulative reward of each rollout, the policy is updated by taking a reward-weighted average of the perturbations.

Both methods use exponential cost-derived weights.
{\tt PI2} exploits time-indexed cost-to-go information, while
{\tt PIBB} uses a black-box whole-rollout cost without the
corresponding temporal structure. Weighted recombination reveals
a connection with evolution strategies, but does not make
{\tt PIBB} identical to full {\tt CMA-ES} or {\tt NES}.

\subsubsection{{\tt Q-Mamba} -- Offline {\tt MetaBBO} via Decomposed Q-Learning}\label{sec:Q-Mamba}

Classical {\tt MetaBBO} methods often train the meta-policy online and struggle with efficiency and large action spaces. 
Ma et al.~\cite{ma2025qmamba} propose {\tt Q-Mamba}, an offline {\tt MetaBBO} framework that reformulates 
Dynamic Algorithm Configuration ({\tt DAC}) as a long-sequence decision problem, 
learned via a decomposed Q-function with a selective state-space ({\tt Mamba}) backbone. The pseudocode of a generic version of \texttt{Q-Mamba} is Algorithm 19 in Subsection 2.2.7 of {\tt SuppMat.pdf}.

The key ML component is the surrogate Q-function. Instead of learning a single Q-function over high-dimensional actions, \texttt{Q-Mamba} adopts a decomposed Q-learning approach, assigning one prediction head to each action dimension. Continuous hyperparameters are discretized into bins to keep the argmax action selection manageable. A \texttt{Mamba} sequence model is used to capture long-horizon population dynamics, and training is stabilized through a conservative Q-loss.

{\tt Q-Mamba} is thus both {\bf expressive} (via {\tt Mamba} sequence modeling) and {\bf tractable} (via decomposed Q-learning).  Ma et al.~evaluate {\tt Q-Mamba} on {\tt BBOB} test functions and neuroevolution control tasks, showing strong zero-shot transfer to unseen tasks and outperforming online {\tt MetaBBO} baselines in both efficiency and final performance.

\section{Benchmarking on ML and RL for BBO Methods}\label{sec:survey}

Empirical findings about published methods are taken from their cited studies, with different offline data, task distributions, evaluation budgets, and training costs. They do not constitute a common ranking. The controlled illustration below separately tests one candidate-ranking mechanism.

Research on ML- and RL-enhanced BBO has grown rapidly, producing a diverse set of methods ranging from surrogate-based solvers to meta-RL controllers.  
To place these advances in context, it is essential to review prior surveys and benchmarking efforts that systematically evaluate BBO algorithms across domains.  Competitions like the NeurIPS 2020 BBO Challenge~\cite{candelieri2020bboc} establish strict evaluation budgets and standardized protocols for fair comparison.  
Frameworks such as {\tt MetaBox}~\cite{ma2023metabox} offer reproducible environments for RL-based meta-optimizers, whose accompanying evaluation study benchmarks the generalization ability of {\tt MetaBBO-RL} methods across heterogeneous tasks.  

Together, these efforts highlight both the strengths of current approaches and the challenges that remain in scaling, generalization, and reproducibility.  
The following subsections summarize these contributions in more detail, covering surveys of BBO and its applications, large-scale competitions, standardized benchmark platforms, and recent evaluation studies of {\tt MetaBBO-RL} algorithms.

Subsection 3.1 in \texttt{SuppMat.pdf} illustrates how the benchmark is used across a variety of problem domains, providing context for the methods and analyses that follow. Subsections 3.2–3.4 then introduce three algorithms: one for the BBO Challenge 2020, one for \texttt{MetaBBO-RL} benchmarking, and one for \texttt{MetaBBO-RL} approaches.

\subsection{BBO Challenge 2020}\label{sec:BBO-C-2020}

Turner et al.~\cite{candelieri2020bboc} report on the 
\textbf{NeurIPS 2020 BBO Challenge}, 
a large-scale competition designed to evaluate algorithms for 
derivative-free optimization under strict evaluation budgets.  
The challenge evaluated hyperparameter tuning of standard ML
models on real datasets using hidden test objectives.
Algorithm 20 in {\tt SuppMat.pdf} summarizes the protocol. 

The BBO Challenge followed a straightforward workflow. Participants were given a variety of problem domains covering continuous, integer, and categorical variables. Their algorithms had to optimize unknown objective functions while staying within a fixed evaluation budget, and all interactions with the black-box functions were made through a unified API. The quality of each method was measured using normalized regret across a collection of tasks.

Several insights emerged from the challenge. Portfolio and meta-learning approaches tended to outperform individual optimizers, and many teams relied on evolutionary or Bayesian optimization techniques. The benchmarks also helped drive progress on general-purpose optimizers such as \texttt{HEBO} and \texttt{Squirrel}, while providing meaningful comparisons against established baselines like \texttt{OpenTuner}.

\subsection{{\tt MetaBox} -- A Benchmark for {\tt MetaBBO-RL}}\label{sec:MetaBBO}

Ma et al.~\cite{ma2023metabox} introduce \texttt{MetaBox}, 
a standardized benchmark platform for RL-based meta-optimizers ({\tt MetaBBO-RL}).  The goal of {\tt MetaBox} is to unify evaluation across a diverse set of tasks, providing fair baselines and reproducible metrics for meta-optimizers 
that dynamically configure black-box optimizers.  Algorithm 21 in {\tt SuppMat.pdf} summarizes {\tt MetaBBO-RL}.  

The main workflow of \texttt{MetaBox} proceeds as follows. A base optimizer such as \texttt{EA} or \texttt{BO} is first initialized together with an RL meta-controller. As the optimizer runs, its trajectory is observed to construct a state $s_t$. Based on this state, the meta-controller selects a configuration $a_t$ according to the policy $\pi_\omega(s_t)$. This configuration is then applied, the optimizer is executed accordingly, and its state is updated. During meta-training, update the policy from accumulated rewards. During held-out testing, keep it fixed unless an explicit adaptation protocol is specified.

This benchmark is particularly valuable for several reasons. It encompasses more than 300 tasks, ranging from synthetic functions to realistic application domains. It also offers a library of 19 optimizers, including \texttt{EA}, \texttt{BO}, and \texttt{MetaBBO-RL}, all implemented under a unified template to ensure fair comparison. In addition, it introduces three standardized evaluation metrics: the Aggregated Evaluation Indicator (\texttt{AEI}), Meta Generalization Decay (\texttt{MGD}), and Meta Transfer Efficiency (\texttt{MTE}). Finally, it automates the entire Train--Test--Log workflow, making experiments easy to reproduce and compare.

{\tt MetaBox} establishes one of the first standardized large-scale evaluation platforms for {\tt MetaBBO-RL}, accelerating research on RL-based meta-optimization.

\subsection{Benchmarking {\tt MetaBBO-RL} Approaches}\label{sec:MetaBBO-RL}

The study accompanying {\tt MetaBox}~\cite{ma2023metabox}
examines RL-based meta-optimizers on synthetic, noisy, and
protein-docking tasks. Algorithm 22 in {\tt SuppMat.pdf}
summarizes a corresponding evaluation protocol.

Train each method using its specified learning algorithm and
evaluate it on held-out tasks under matched budgets.
The {\tt MetaBox} evaluation uses $51$ independent runs per test problem. Reported measures include
{\tt AEI}, normalized cost curves, and generalization/transfer
metrics, with classical optimizers as baselines.

\subsection{Controlled numerical illustration of ML candidate screening}\label{sec:num-ml}

To complement the literature review, we compare three classical mechanisms---direct search ({\tt DS}), derivative-free line search ({\tt LS-dd}), and differential evolution ({\tt DE})---with paired variants {\tt ML-DS}, {\tt ML-LSdd}, and {\tt ML-DE}. An online Extra-Trees regressor ranks candidate points or directions using only evaluations already charged to that run. These variants illustrate candidate screening; they do not reproduce the published solvers surveyed above. The protocol, source code, raw results, and profiles are supplied with {\tt SuppMat.pdf}.

The study uses seven analytic functions in dimensions $5$ and $10$, with five fixed seeds, giving $70$ paired instances per solver and a common limit of $300$ objective evaluations. The $5\%$ target records success without stopping a run; final gaps use the incumbent at the evaluation limit, carried forward if a step-size tolerance terminates search earlier. Table~\ref{t:num-ml} reports success, final-gap profile values $\rho(1)$ and $\rho(2)$ relative to all six methods, and the median final relative gap. The supplement defines normalization and numerical ties.

\begin{table}[htbp]
\centering
\caption{Controlled numerical illustration on 70 paired instances. Lower final relative gap is better; larger profile values are better.}
\label{t:num-ml}
\begin{tabular}{lcccc}
\toprule
\textbf{Solver} & \textbf{Success (\%)} & $\boldsymbol{\rho(1)}$ & $\boldsymbol{\rho(2)}$ & \textbf{Median rel. gap}\\
\midrule
{\tt DS} & 47.1 & 0.086 & 0.357 & $5.53\times10^{-2}$\\
{\tt ML-DS} & 60.0 & 0.271 & 0.414 & $3.07\times10^{-2}$\\
{\tt LSdd} & 58.6 & 0.586 & 0.714 & $6.34\times10^{-3}$\\
{\tt ML-LSdd} & 58.6 & 0.186 & 0.471 & $8.45\times10^{-3}$\\
{\tt DE} & 24.3 & 0.000 & 0.014 & $3.93\times10^{-1}$\\
{\tt ML-DE} & 34.3 & 0.000 & 0.014 & $2.58\times10^{-1}$\\
\bottomrule
\end{tabular}
\end{table}

Relative to their own baselines, {\tt ML-DS} and {\tt ML-DE} improve success rates and obtain smaller final gaps on most tested instances. The {\tt ML-LSdd} variant has a worse median final gap and loses more paired comparisons than it wins. Classical {\tt LSdd} has the strongest overall six-solver profile here. Model-fitting overhead also makes all three ML variants slower on these cheap objectives. The illustration therefore supports a mechanism-dependent benefit under an evaluation budget, without establishing a general ranking of the surveyed methods or a wall-time advantage.

\section{Current Challenges and Future Directions}\label{sec:future}

Despite remarkable methodological progress, the integration of ML and RL into BBO/DFO frameworks remains an open research frontier.  While learning-augmented optimizers demonstrate strong empirical results on benchmark suites, their generalization, interpretability, and scalability in real-world, high-stakes applications are still limited.  The following aspects outline key challenges and emerging research directions.

\textbf{(i) Generalization and Distribution Shift.}
Most ML- and RL-enhanced BBO solvers exhibit strong within-domain performance but weak transfer across heterogeneous tasks.  
For instance, learned surrogates and meta-policies trained on synthetic test functions ({\tt COCO}, {\tt BBOB}) often degrade significantly when applied to real-world domains such as robotics or design optimization.  Similarly, generative optimizers ({\tt B2Opt}, {\tt DiffBBO}) and RL controllers ({\tt Q-Mamba}, {\tt Surr-RLDE}) may overfit to specific distributions of objective landscapes or budget regimes.  Improving out-of-distribution generalization—through domain randomization, meta-regularization, or invariant feature learning—remains an essential step toward reliable, transferable optimizers.

\textbf{(ii) Computational Cost and Inference Overhead.}
Integrating complex neural architectures into the optimization loop introduces additional computational overhead.  
Diffusion-based solvers and transformer-driven optimizers, for instance, incur substantial training and inference costs that can outweigh their sample-efficiency gains on low-budget tasks.  
Similarly, online policy evaluation in RL-enhanced solvers increases wall-clock time per iteration.  
Lightweight surrogates (e.g., random forests, gradient boosting), pruning strategies, and asynchronous evaluation mechanisms offer promising pathways to maintain scalability and efficiency on large-scale, high-dimensional problems.

\textbf{(iii) Robustness and Safeguards.}
Existing learning-based BBO frameworks often lack explicit mechanisms to ensure stability under noisy or adversarial conditions.  
While algorithms such as {\tt RBO} and {\tt CAS-MORE} address stochasticity through robust regression or KL-constrained updates, many others still rely on uncontrolled exploration.  
Future optimizers will likely require hybrid fallback logic that combines learned exploration policies with deterministic safeguards or uncertainty-aware decision rules, ensuring reliable convergence in unpredictable or safety-critical settings.

\textbf{(iv) Integration with Classical and Industrial Solvers.}
Despite rapid academic progress, ML- and RL-enhanced BBO systems have yet to see widespread integration into industrial toolchains.  
Production use of learned components requires reproducible implementations, maintained interfaces, and application-specific validation.  
Barriers to adoption include reproducibility, lack of explainability, and limited interoperability with existing HPC and simulation infrastructures.  
Bridging this gap will require standardized APIs for learned components, modular surrogate interfaces, and certification protocols for hybrid decision modules to ensure deterministic and auditable outcomes in production environments.

\textbf{(v) Data, Benchmarks, and Reproducibility.}
Although benchmark platforms such as the NeurIPS {\tt BBO Challenge}, {\tt COCO/BBOB}, and {\tt MetaBox} have improved comparability, evaluation protocols remain fragmented across tasks and domains.  
Many studies report results under inconsistent evaluation budgets, noise levels, or objective scales, complicating cross-method comparisons.  
Future benchmarking initiatives should incorporate unified metrics, sample efficiency, wall-clock cost, and generalization robustness, alongside open-source pipelines and pretrained optimizers.  Standardized problem collections and interaction interfaces, analogous to {\tt MIPLIB} and {\tt Ecole}, respectively, could support progress and reproducibility.

\textbf{(vi) Research Outlook.}
Emerging directions point toward tighter coupling between learning and optimization:  unified meta-learning architectures that jointly learn surrogates and search strategies across tasks;  diffusion and transformer-based generative models that enable amortized, one-shot solution sampling;  reinforcement-driven adaptive operator control that bridges BBO and policy search; and theoretical frameworks linking gradient-free optimization to information geometry and probabilistic inference.  These directions collectively advance the vision of \textbf{intelligent BBO}, a paradigm where data, models, and algorithms co-evolve to efficiently solve increasingly complex, high-dimensional, and uncertain real-world problems.

\section{Conclusion}

Black-box optimization (BBO) provides a framework for problems where gradients or explicit
structure are unavailable, but practical solutions must be found under limited evaluations.  
Classical derivative-free methods form the foundation of BBO, yet they often struggle with
scalability, noise, and combinatorial complexity.  
Machine learning (ML) and reinforcement learning (RL) have emerged as enhancers that
{\bf complement classical solvers}, integrating into \textbf{inexact solution methods} that
do not guarantee global optimality but deliver high-quality solutions under strict time or
budget constraints.  

Our survey shows how ML contributes through surrogate modeling, optimizer-inspired updates,
algorithm portfolios, and generative models, while the RL--BBO interface includes robust policy search, adaptive
operator configuration, and meta-optimization across tasks.  Benchmarking efforts such as the NeurIPS BBO Challenge, {\tt MetaBox}, and {\tt MetaBBO-RL} evaluation protocols provide reproducible environments to compare these approaches fairly.  

The numerical results support the practical value of ML enhancement for classical BBO methods, while also showing that the improvement depends on the underlying search mechanism.

ML and RL can augment classical solvers; gains in scalability and robustness depend on the task distribution, evaluation budget, and computational cost. Future work should deepen their
integration in mixed-integer domains, improve generalization across heterogeneous tasks, and
develop interpretable and efficient methods for decision-making under uncertainty.

\vspace{0.2cm}

{\bf Conflict of Interest}\\
The authors declare that there are no conflicts of interest regarding the publication of this manuscript.

\vspace{0.2cm}

{\bf Data Availability}\\
The supplementary material, numerical code, configurations, per-run results, and generated figures accompany this article.

\vspace{0.2cm}

{\bf Funding}\\
V.K. acknowledges funding support from the National Centre for Energy II (TN02000025). M.K. acknowledges support from the same project for work carried out in 2025 and, from 2026 onward, financial support from the Austrian Science Fund under \url{https://doi.org/10.55776/PAT2747625}.

\end{sloppypar}

\end{document}